\documentclass[10pt]{article}
\usepackage[preprint]{tmlr}
\usepackage{enumitem}

\usepackage{amsmath,amssymb,amsthm}
\usepackage{bm}
\usepackage{booktabs}
\usepackage{multirow}
\usepackage{graphicx}
\usepackage{subcaption}
\usepackage{xr-hyper}
\usepackage{hyperref}
\usepackage{xcolor}
\usepackage[normalem]{ulem}

\newif\ifrevisionmarkup
\revisionmarkupfalse

\usepackage{environ}

\NewEnviron{revdelblock}{\ifrevisionmarkup{\color{red}\BODY}\fi}
\usepackage[most]{tcolorbox}
\usepackage{listings}
\usepackage{float}
\usepackage{placeins}
\usepackage{adjustbox}
\usepackage{array}

\title{Correction and Corruption: A Two-Rate View of Error Flow in LLM Protocols}
\author{
\name Fernando Reitich \email fernando.reitich@imens.tech \\
\addr Imens, LLC \\
\addr Weston, Florida, USA
}
\date{August 10, 2026}

\newcommand{\suppref}[1]{\ref*{#1}}

\begin{document}
\maketitle

\begin{abstract}
Large language models increasingly operate in protocols containing multiple
model calls, yet added calls are usually evaluated only by their net effect on
success. That summary cannot distinguish two opposing changes: correcting
initially unsuccessful outputs and corrupting initially successful ones. We
develop a paired audit that records success before and after a specified
operation on the same task instances under one common binary rule. Correction
and corruption rates then give an exact accounting of the net change: gains
come from corrected failures and losses from corrupted successes. The
break-even correction requirement rises sharply with baseline success, so the
same correction and corruption behavior can improve a moderate-baseline
population but harm a high-baseline one. Applied to published GPT-4 GSM8K
results, the framework bounds the within-sample correction and corruption
counts from aggregate accuracies.
We then ask three questions for protocol and agent design: whether calibration
estimates predict unobserved target outcomes, how rates change with information
supplied to an operation, and whether measurements of successive operations
combine. For prediction, calibration estimates track post-operation accuracy
on a disjoint sample from the same generator. Under controlled reweighting of
generator-defined task groups, pooled estimates can fail, while group-specific
estimates reduce average prediction error. On GSM8K, readily observable
features of the problem statement capture only limited variation in correction
and corruption at the available sample sizes, so the group-level application
rule remains exploratory. For input sensitivity, reordering a fixed
four-candidate set changes both rates, with the accuracy effect depending on
whether a correct alternative is available; on MBPP, adding a helper artifact
raises corruptions among 357 initially passing programs from \(28\) to \(100\).
For composition, direct and composed transition estimates are close on average
within samples; in held-out comparisons, weighting the initial-success rows by
their support substantially reduces the discrepancy.
Correction and corruption are therefore measurements of a specified
operation, task population, input configuration, and success rule---not fixed
properties of a model. For protocol and agent design, paired outcomes should
be logged at each operation so the resulting rates can inform comparisons of
input choices, application decisions, and tests of composition. Each use
requires representative calibration data and held-out evaluation; application
decisions must also account for the cost of the added call.
\end{abstract}

\section{Introduction}
\label{sec:introduction}

Large language models increasingly operate within multi-call systems rather than as isolated predictors. We use \emph{protocol} for a structured sequence of model calls, including both fixed pipelines and adaptive agent workflows. Yet the effect of an added call is usually summarized only by the change in aggregate success.

That summary can conceal two opposing changes. As we detail in Section~\ref{sec:paired_outcomes}, on the same 500 arithmetic problems, supplying a Mistral worked solution to Llama~3.2 corrected 32 initially wrong answers but corrupted 114 initially correct answers, reducing accuracy from \(0.838\) to \(0.674\). Reversing the model roles corrected 263 initially wrong answers and corrupted only 10 initially correct answers, increasing accuracy from \(0.346\) to \(0.852\). The two baselines differ because reversing the roles changes the model whose independent answer is evaluated. Aggregate accuracy shows the net loss in the first direction and the net gain after the roles are reversed; the paired outcomes reveal the corrections and corruptions underlying each change.

This paper develops a paired audit of how success changes when one specified model call is applied to an existing response for each task. The initial response may come from the same model or from another model. Success is evaluated before and after the call by the same pre-specified binary rule: for example, agreement with a reference answer, passage of executable tests, successful execution, or satisfaction of a fixed constraint. A unique ground-truth output is not required. An initially unsuccessful output that becomes successful is a \emph{correction}; an initially successful output that becomes unsuccessful is a \emph{corruption}. Their rates \(c\) and \(\gamma\) are the corresponding conditional proportions, and their weighted difference is exactly the observed change in success rate.

We call one model invocation within a larger system a \emph{protocol operation}. An operation is specified by its model, instruction, decoding rule, and required output, and receives task-specific inputs such as the current answer, feedback, candidate outputs, retrieved evidence, or tool results. Changing the specification defines a different operation; holding it fixed while changing what is presented tests dependence on the operation's inputs. Correction and corruption therefore describe a specified operation, task population, input configuration, and success rule---not a model in isolation.

On a population for which all paired outcomes are observed, this accounting is complete. Using the rates beyond that retrospective description requires at least one of three moves: applying the same operation to unobserved tasks, changing a controllable feature of the operation or its inputs, or combining it with other calls. These yield three basic questions for protocol design. First, can rates estimated from calibration data predict the effect of the same operation on a target sample? Different task groups can have different rates, so pooled estimates may fail when calibration and target populations contain different mixtures. Second, how do the rates change under controllable choices in the operation or its inputs? Without understanding this dependence, measurements for one call cannot guide the design of a nearby alternative. Third, when calls are chained, can their separately measured rates be combined? Composition requires the information retained between calls to be adequate for predicting the next operation. Together, these questions ask whether a local measurement can be reused, deliberately changed, and incorporated into a larger protocol or agent.

The experiments herein follow this progression. Controlled arithmetic tasks use structural depth to define groups with different rates and to test prediction under matched holdout and changed population composition; \(2\times2\) linear systems provide a second exact-scoring family. GSM8K asks whether readily observable task features separate groups with meaningfully different rates. A judge-based selection experiment holds a realized candidate set fixed and changes only presentation order. Reordering changes which candidate is selected and the resulting correction and corruption rates; it can hurt when diversity offers no genuine alternative and help when a correct alternative is present. A targeted MBPP audit extends the analysis to a setting in which correctness is defined by unit tests. It measures the effect of supplying a helper artifact produced by another model. Adding the artifact increases the number of corrupted programs among 357 initially passing programs from \(28\) to \(100\), while the task, initial program, revision model, decoding settings, core instruction, and tests remain fixed. Because the experiment conditions on initially passing programs, it isolates corruption by construction. Finally, a two-operation synthetic stack tests whether measurements from adjacent operations combine when intermediate success is used as the state summary.

The empirical evidence is deliberately bounded. The principal comparisons use open-weight models from the Llama 3.2, Mistral, and Qwen 2.5 families. Broader domains and substantially different capability levels remain outside our direct experiments.

Prior work is relevant in two respects. Research on self-refinement, critique, verification, and feedback-assisted reasoning establishes both the prevalence of additional calls and the possibility that they improve some outputs while destabilizing others \citep{madaan2023selfrefine,huang2023llms,yang2025confidencecritique,yang2025scalingtheory}. Section~\ref{sec:reported_accuracy_bounds} uses the results reported by \citet{huang2023llms} to show what aggregate before-and-after accuracies can and cannot determine about correction and corruption when paired outcomes are unreported. \citet{yang2025confidencecritique,yang2025scalingtheory} provide the closest prior formal precedents. They distinguish the repair of initially wrong answers from the retention of initially correct answers and extend this analysis to multiple rounds of self-correction. Independent contemporaneous work by \citet{liu2026feedbackcontrol} defines the Error Correction Rate and Error Introduction Rate, which coincide with \(c\) and \(\gamma\), respectively, and incorporates them into a two-state Markov model of repeated self-revision. Their experiments include proprietary systems not tested directly here. Together, these studies provide evidence across additional models and task settings that the correction--corruption distinction is not confined to our experimental setting.

The present paper treats self-revision as one instance of a broader paired audit of protocol operations and addresses the three questions posed above for design use: prediction on unobserved tasks, dependence on the operation and its inputs, and composition across adjacent, potentially different operations.

The rest of the paper is organized as follows.
Section~\ref{sec:paired_measurement} defines the paired audit and its limits
of identification; Sections~\ref{sec:prediction}--\ref{sec:successive_steps}
study prediction, dependence on the operation and its inputs, and
composition; Section~\ref{sec:discussion} discusses scope and limitations;
and Section~\ref{sec:conclusion} concludes.

\section{Paired Measurement of Correction and Corruption}
\label{sec:paired_measurement}

\subsection{Paired Outcomes and Exact Accounting}
\label{sec:paired_outcomes}

Consider a fixed population of \(N\) task instances and one specified protocol operation. Let \(E_0\in\{0,1\}\) indicate whether the output before the operation is successful, and let \(E_1\in\{0,1\}\) indicate whether the output after the operation is successful under the same rule. Write
\begin{equation}
C_{ij}
=
\#\{x:E_0(x)=i,\ E_1(x)=j\},
\qquad i,j\in\{0,1\}.
\label{eq:paired_counts}
\end{equation}

The four cells have direct meanings. \(C_{01}\) counts corrections, \(C_{10}\) counts corruptions, \(C_{00}\) counts outputs that remain unsuccessful, and \(C_{11}\) counts outputs that remain successful. Let
\[
p_0=\Pr(E_0=1),
\qquad
p_1=\Pr(E_1=1),
\]
and define the correction and corruption rates
\begin{equation}
c
=
\Pr(E_1=1\mid E_0=0)
=
\frac{C_{01}}{C_{00}+C_{01}},
\qquad
\gamma
=
\Pr(E_1=0\mid E_0=1)
=
\frac{C_{10}}{C_{10}+C_{11}}.
\label{eq:correction_corruption_rates}
\end{equation}
These rates are defined provided there is at least one initially unsuccessful
and one initially successful instance, respectively; otherwise the corresponding
rate is undefined.

On this fixed paired population,
\begin{equation}
p_1-p_0=(1-p_0)c-p_0\gamma.
\label{eq:paired_accounting}
\end{equation}
When a conditioning set is empty, its weighted contribution in
Equation~\eqref{eq:paired_accounting} is taken to be zero.
Equation~\eqref{eq:paired_accounting} is an exact accounting identity, not an empirical model. The first term is the fraction of all items corrected by the operation; the second is the fraction corrupted. In counts, improvement is simply \(C_{01}>C_{10}\). When \(0<p_0<1\), the same condition expressed through conditional rates is
\begin{equation}
c>\frac{p_0}{1-p_0}\gamma.
\label{eq:break_even}
\end{equation}
Replacing the strict inequality by equality gives the \emph{break-even boundary} between net
improvement and net loss.

The break-even requirement also depends strongly on the baseline success rate.
The multiplier \(p_0/(1-p_0)\) rises from \(3\) at \(p_0=0.75\) to \(19\) at
\(p_0=0.95\). Thus, holding the corruption rate fixed, the correction rate
required for break-even is more than six times larger at the higher baseline.
Because \(c\leq1\), at a baseline of \(0.95\) no correction rate can offset a
corruption rate above \(1/19\). The same correction and corruption behavior can
therefore improve a moderate-baseline population but harm a high-baseline one.
Net changes observed at different baselines are not directly comparable without
the two underlying rates.

Table~\ref{tab:paired_alt_example} reports the two directions of the operation described in the Introduction, which we call \emph{helper-transcript conditioning}. In each direction, one model supplies a worked solution and the other re-solves the same 500 arithmetic problems while conditioned on it. The two baselines differ because, in each row, \(p_0\) is the re-solving model's independent accuracy on these problems. All quantities are raw proportions computed directly from the displayed paired counts, and they satisfy Equation~\eqref{eq:paired_accounting} exactly. The role reversal changes both flows, not merely the baseline accuracy. The rates therefore describe the specified operation, direction, and task population; they are not intrinsic constants of either model.

\begin{table}[htbp]
\centering
\caption{\textbf{Paired outcomes when each model, in turn, supplies a worked solution to the other on the same 500 arithmetic problems (\(N=500\) in each row).} The table reports raw counts and raw rates. Supplying a Mistral worked solution to Llama~3.2 corrects 32 of 81 initially wrong answers but corrupts 114 of 419 initially correct answers, reducing accuracy from \(0.838\) to \(0.674\). Reversing the roles corrects 263 of 327 initially wrong answers and corrupts 10 of 173 initially correct answers, increasing accuracy from \(0.346\) to \(0.852\). The contrast is descriptive of these specified model roles, prompts, and task population.}
\label{tab:paired_alt_example}
\small
\begin{adjustbox}{max width=\linewidth}
\begin{tabular}{@{}lrrrrrrrr@{}}
\toprule
Worked solution supplied to target
& \(C_{00}\)
& \(C_{01}\)
& \(C_{10}\)
& \(C_{11}\)
& \(p_0\)
& \(p_1\)
& \(c\)
& \(\gamma\) \\
\midrule
Mistral supplied to Llama~3.2
& 49 & 32 & 114 & 305 & 0.838 & 0.674 & 0.395 & 0.272 \\
Llama~3.2 supplied to Mistral
& 64 & 263 & 10 & 163 & 0.346 & 0.852 & 0.804 & 0.058 \\
\bottomrule
\end{tabular}
\end{adjustbox}
\end{table}

Equation~\eqref{eq:paired_accounting} always explains the observed change on the population from which its four cells are computed. Section~\ref{sec:prediction} turns to the distinct prospective use of rates estimated from calibration data.

\subsection{Scope of the Paired Audit}
\label{sec:paired_scope}

The paired audit applies to a specified before-and-after operation, not to a model in isolation. Four conditions define the measurement.

First, the same task instances must be observed before and after the operation. Without pairing, the aggregate success rates may constrain the possible flows but do not reveal which items changed.

Second, the two checkpoints must use one common binary success rule. Numeric exact match is one instance. Unit-test passage, successful execution, satisfaction of a fixed constraint, or a predetermined adjudication rule may serve the same role. The audit therefore does not require a unique reference output, but it does require an auditable criterion.

Third, the operation must be specified. A helper transcript followed by re-solving, a verify-and-fix call, selection from a fixed candidate set, and program revision with additional context are different operations. Each can be audited when the pre- and post-operation bits represent success of the output under the same rule. Their rates remain tied to the task population, the operation as specified, the input configuration---including the source of supplied information---and the success rule.

Fourth, both bits must encode the same outcome property: whether the system
output is successful at the corresponding checkpoint. Sharing a rule is not
enough if the bits score different quantities. Verdict-only verification
illustrates this distinction. The first bit records whether the candidate
answer is correct, whereas the second records whether the judge labels it
correct. The resulting quantities are a false-positive rate, a false-negative
rate, and an acceptance rate, not correction, corruption, and final-answer
accuracy. We therefore treat verdict-only verification as a classification
diagnostic rather than place it on the same empirical axes as
answer-transforming operations.

These requirements delimit what the audit establishes. It separates the observed corrections and corruptions underlying a net change. When two operations or input conditions are compared on the same pre-operation outputs, a difference in net effect can arise from changed correction, changed corruption, or both. It does not identify why a model changed its output, nor does it measure general model competence. The relevant object is the complete specified setting: task population, protocol operation, input configuration, and success rule.

The same distinction governs the empirical claims that follow. Methodologically, any operation satisfying the paired measurement conditions can be audited. Empirically, the evidence reported here comes from selected operations on controlled mathematics and GSM8K and from a one-sided MBPP corruption comparison.

\subsection{What Aggregate Accuracies Do and Do Not Determine}
\label{sec:reported_accuracy_bounds}

Published evaluations often report accuracy before and after revision without
reporting which individual answers changed. Those totals determine the net
difference between corrections and corruptions, but not the two counts
separately. The paired framework can nevertheless bound the within-sample
flows compatible with the reported totals, and known protocol rules can
sometimes determine them more sharply. These are constraints on the fixed
reported sample, not estimates intended to predict new problems.

\citet{huang2023llms} report that GPT-4 accuracy on a 200-item GSM8K sample
falls from \(191/200\) to \(183/200\) after one round of intrinsic
self-correction. If \(C_{01}\) answers are corrected and \(C_{10}\) are
corrupted, the eight-item loss implies
\[
C_{10}-C_{01}=8.
\]
Only nine answers were initially wrong, so \(C_{01}\) can range from \(0\) to
\(9\), while \(C_{10}\) ranges from \(8\) to \(17\). The published totals
therefore constrain the corruption count but leave the correction rate
\(c=C_{01}/9\) entirely undetermined within the sample.

Huang et al.\ also report a ground-truth-guided procedure, which they call
oracle-guided self-correction, whose accuracy rises from \(191/200\) to
\(195/200\). The aggregate accuracies alone allow several paired tables, but
in this procedure answers already known to be correct are not revised, so
initially correct answers cannot be corrupted: \(C_{10}=0\). The four-item gain
must therefore consist of exactly four corrections, \(C_{01}=4\).

This reanalysis does not change Huang et al.'s aggregate conclusions. It adds
a flow-level interpretation: aggregate results can bound the unreported
correction and corruption counts, and protocol structure can sometimes
determine them, but paired logs are needed when no such restriction is
available.
\section{Prediction Using Correction and Corruption Rates}
\label{sec:prediction}

Section~\ref{sec:paired_measurement} established exact accounting once the paired outcomes on a
population are observed. This section asks the first of the design questions posed in
Section~\ref{sec:introduction}: whether correction and corruption rates estimated from calibration
data can be used prospectively, before the target outcomes are known. For a specified operation, task
population, input configuration, and success rule, \(c\) and \(\gamma\) are population quantities,
while calibration data provide estimates \(\hat c\) and \(\hat\gamma\).

We proceed in four steps. Section~\ref{sec:structural_depth} uses a matched holdout from the same task
distribution to assess finite-sample prediction. Section~\ref{sec:population_composition} changes only the
depth mixture; because correction and corruption rates can differ across depth groups, pooled calibration
estimates can then give inaccurate predictions for the reweighted population. Section~\ref{sec:gsm8k_features}
moves to GSM8K and asks whether observable features of the problem statement define groups with
meaningfully different rates. Section~\ref{sec:selective_application} asks whether estimated rates can
support an application decision.

The controlled analysis begins with helper-transcript conditioning. For an instance $x$, a helper
model $A$ produces a worked solution $r_A(x)$, and the target model $B$ solves the same instance while
conditioned on that solution. We write \(A\to B\) for this ordered role assignment: \(A\) supplies the
helper transcript and \(B\) produces the post-operation answer. The two rows of Table~\ref{tab:paired_alt_example}
are therefore the assignments mistral\(\to\)llama3.2 and llama3.2\(\to\)mistral, respectively.
The holdout analysis below also includes llama3.2\(\to\)llama3.2 as a same-model comparison.
Because $B$ is not asked to critique or verify the helper, the requested action is held fixed while only
the information supplied to the call
changes. As in Section~\ref{sec:paired_outcomes}, $E_0$ records success of $B$'s independent answer and
$E_1$ records success after helper-transcript conditioning, under the same rule.

The synthetic corpus supplies both exact outcomes and a known grouping variable.
For each family, the fixed randomized generation procedure induces its task
distribution. Task instances are sampled by seeded pseudorandom generators and then
scored exactly. The first family is depth-stratified arithmetic with structural depth
\(d\in\{1,\dots,5\}\): integer operands and operators from
\(\{+,-,\times,\div\}\) are sampled under constraints on operand range and
well-formedness, including avoidance of degenerate cases such as division by
zero, and recursively composed to the requested depth. The second family samples
\(2\times2\) linear systems with integer coefficients. For both families, the generation seed fixes the pseudorandom stream
and the task ID identifies an instance within that stream; the same seed--task-ID
pair therefore reproduces the same problem, while different seeds provide
separate reproducible samples from the same family-specific task distribution.

Each of the three generation seeds (123, 124, 125) contributes 600 items: 100 at
each of the five arithmetic depths and 100 linear systems. The 500 arithmetic
items generated under seed 123 are the sample reported in
Table~\ref{tab:paired_alt_example}. The prediction analyses in
Sections~\ref{sec:structural_depth}--\ref{sec:population_composition} use the five
arithmetic depth groups; the composition audit in
Section~\ref{sec:successive_steps} uses all six groups, with the linear systems
providing a structurally distinct exact-scoring group. 
The baseline solve and helper-transcript conditioning used in
Sections~\ref{sec:structural_depth}--\ref{sec:population_composition}, and the
verify-and-fix operation added in Section~\ref{sec:successive_steps}, all use
deterministic decoding. Variation across seeds in these analyses therefore
comes from the generated task samples rather than from model sampling.

An output is successful when it matches the generator's ground-truth answer
exactly after deterministic normalization: a shared formatter reduces the
free-form transcript to a single-line output, which a parser maps to a canonical
form (integers, reduced rationals, and terminating decimals mapped to rationals
where unambiguous).

Within the arithmetic family, depth is supplied by the generator rather than inferred. It therefore
provides a grouping variable without requiring any claim that depth measures difficulty in general,
and lets us compare rates estimated within each depth group with a single pooled estimate formed after
combining all depth groups. We call the relative proportions of the depth groups in an evaluation
population its depth mixture. Let \(X\) denote a task instance drawn from the evaluation population,
and let
\(s(X)\in\{1,\dots,5\}\) denote its generator-supplied depth. We write
\[
c(d)=\Pr(E_1=1\mid E_0=0,\ s(X)=d),\qquad
\gamma(d)=\Pr(E_1=0\mid E_0=1,\ s(X)=d).
\]
Hats denote sample estimates. We use the Jeffreys-smoothed estimates
\[
\hat c=\frac{C_{01}+1/2}{C_{00}+C_{01}+1},
\qquad
\hat\gamma=\frac{C_{10}+1/2}{C_{10}+C_{11}+1},
\]
with the counts taken in the relevant pool or depth group. Jeffreys smoothing stabilizes estimates when
the conditioning set is small, while shrinking them toward \(1/2\). If a conditioning set is empty,
we report the corresponding rate as undefined rather than the smoothed value \(1/2\). The quantities
\(\hat p_0\) and \(\hat p_1\) denote empirical success proportions, and \(\hat p_t(d)\)
denotes the corresponding empirical proportion at checkpoint \(t\) within depth group \(d\). The raw counts and raw rates for the running example
remain those of
Table~\ref{tab:paired_alt_example}.

\subsection{Prediction on Held-Out Samples from the Same Generator}
\label{sec:structural_depth}

We first use a matched holdout as a control for the population-composition experiment in
Section~\ref{sec:population_composition}. We fit on two generation seeds and predict the third,
leaving the depth mixture unchanged, so that calibration and target samples share the same task
distribution, depth definitions, models, prompts, decoding, and success rule. This design holds the
population quantities \(c(d)\) and \(\gamma(d)\) fixed and asks how accurately their calibration estimates
predict the realized held-out outcomes at the available sample sizes. It is therefore a finite-sample
prediction check; the deliberate change of depth mixture enters only in
Section~\ref{sec:population_composition}.

\begin{figure}[htbp]
  \centering
  \includegraphics[width=0.98\linewidth]{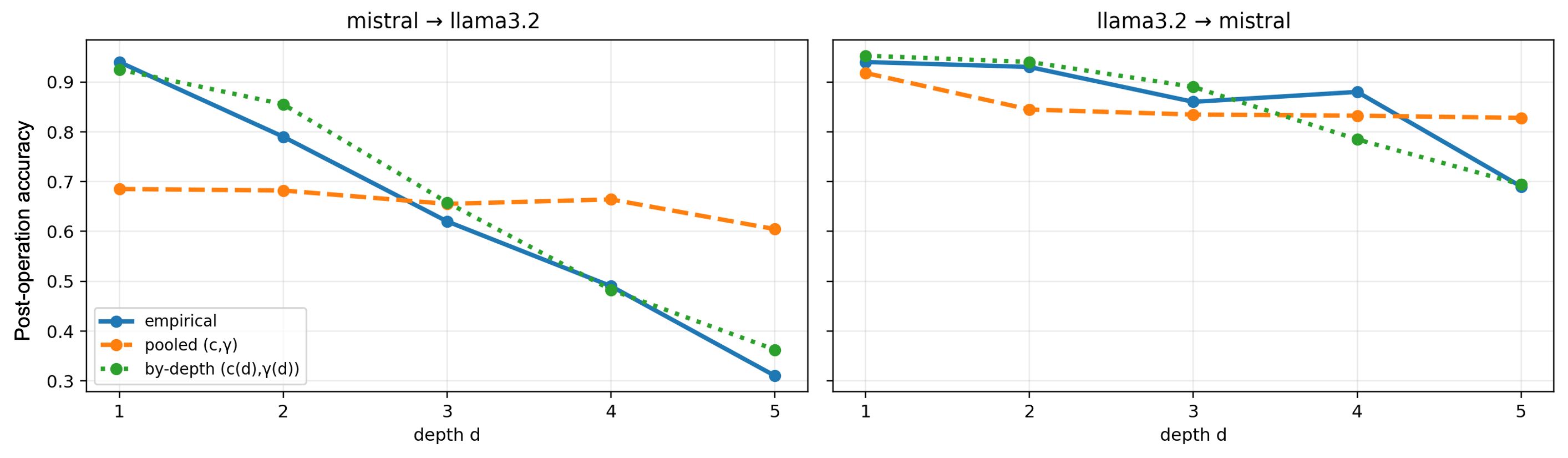}

\caption{\textbf{Held-out depth profiles for helper-transcript conditioning.}
The curves use calibration seeds 123/124 and held-out seed 125. Left:
\texttt{mistral}$\to$\texttt{llama3.2}; right:
\texttt{llama3.2}$\to$\texttt{mistral}. For each depth bin $d$, the solid
curve is empirical post-operation accuracy $\hat p_1(d)$; the dashed curve
predicts it from held-out $\hat p_0(d)$ using one pooled
$(\hat c,\hat\gamma)$ fit; and the dotted curve uses
$(\hat c(d),\hat\gamma(d))$. Using a single pooled pair removes depth-specific
variation in \(c\) and \(\gamma\); any remaining variation in the pooled prediction across depth comes
from the held-out $\hat p_0(d)$, and in this holdout that leaves the pooled profile close to flat. The
empirical profile falls overall with depth and the depth-conditioned prediction follows that trend. Calibration and held-out samples share the generator,
operation, depth definitions, and depth mixture.}
\label{fig:seed-holdout-depth-profiles}
\end{figure}

Across the three held-out folds, mean absolute error for the depth-conditioned predictor over the five depth-specific post-operation
accuracy predictions ranges from $0.009$--$0.015$ for the same-model pair and from $0.027$--$0.051$
across the two directed cross-model pairs, with the largest per-depth absolute error over all held-out
seeds and directed model pairs about \(0.144\). The matched holdout also reveals depth-specific
variation before any mixture change. Figure~\ref{fig:seed-holdout-depth-profiles} shows where the
pooled and depth-conditioned predictors part company. The pooled predictor uses the same
$(\hat c,\hat\gamma)$ in every depth group but evaluates them at the held-out baseline accuracy
$\hat p_0(d)$; its remaining variation across depth therefore comes from $\hat p_0(d)$ rather than
from depth-specific rates, and in this holdout that leaves the pooled profile close to flat. The empirical
profile falls overall with depth, and the depth-conditioned prediction follows that trend. Thus depth-specific variation is visible even when calibration and target
samples share the same depth mixture.

Depth conditioning also divides the relevant denominators: \(c(d)\) uses only initially unsuccessful
items at depth \(d\), and \(\gamma(d)\) only initially successful ones. Thin conditional denominators
can therefore increase uncertainty. Subsampling and conditional-support diagnostics are reported in
the Supplementary Material.

This matched holdout is a prerequisite and control for the reweighting test that follows. At the
sample sizes used here, calibration estimates predict a disjoint sample from the same generator-defined
task distribution with the errors reported above, while Figure~\ref{fig:seed-holdout-depth-profiles}
shows depth-specific variation that a pooled pair cannot represent. The result does not establish
transfer to a different task family, model, operation specification, or input condition, and the estimates
are not constants of either model.

\subsection{Prediction under Changed Population Composition}
\label{sec:population_composition}
\label{sec:synthetic_mixture}

Section~\ref{sec:structural_depth} held the depth mixture fixed. The remaining question is what
happens when the operation and the behavior within each depth remain fixed, but the target population
contains a different mix of depths. To isolate that change, we fit the rates on two seeds and evaluate predictions
on the third. For each held-out seed, we keep the fitted rates and all task outcomes fixed and vary only
the evaluation weights \(w(d)\) over the five depth groups. We compare the original equal-depth mixture
with five counterfactual mixtures, each concentrated on one depth. No additional model calls are made.
A pooled predictor uses one pair \((\hat c,\hat\gamma)\), whereas a depth-conditioned predictor uses
\(\bigl(\hat c(d),\hat\gamma(d)\bigr)\); both predict held-out post-operation accuracy from held-out
baseline accuracy through Equation~\eqref{eq:paired_accounting}. 

Table~\ref{tab:mixture_shift_mae} summarizes the prediction errors for llama3.2\(\to\)mistral across the three held-out seeds. Mean absolute error in aggregate
post-operation accuracy under the original equal-depth mixture is \(0.0078\) for the pooled predictor
and \(0.0084\) for the depth-conditioned predictor. Across the five single-depth mixtures and three
held-out seeds, the corresponding mean absolute error is \(0.0711\) versus \(0.0375\). Taking, for each held-out
seed, the largest absolute error across the five single-depth mixtures and then averaging over seeds
gives \(0.1408\) versus \(0.0990\). Thus the two predictors perform similarly under the original
mixture, while depth conditioning roughly halves average prediction error and lowers the average
per-seed worst-depth error after reweighting.

\begin{table}[htbp]
\centering
\caption{\textbf{Prediction error under controlled reweighting of the depth mixture.}
Mean absolute error in aggregate post-operation accuracy for helper-transcript conditioning in the
direction \texttt{llama3.2}$\to$\texttt{mistral}, summarized over the three choices of held-out seed.
The original mixture uses equal weight across depths; each single-depth mixture concentrates all
evaluation weight on one depth. Conditioning lowers average error and the average per-seed worst-depth
error under reweighting. The generator, operation, and per-depth outcomes remain fixed; only the
evaluation weights change.}
\label{tab:mixture_shift_mae}
\begin{tabular}{@{}lrr@{}}
\toprule
Evaluation composition & Pooled MAE & Depth-conditioned MAE \\
\midrule
Original equal-depth mixture & 0.0078 & 0.0084 \\
Single-depth mixtures, mean & 0.0711 & 0.0375 \\
Worst single depth, averaged over held-out seeds & 0.1408 & 0.0990 \\
\bottomrule
\end{tabular}
\end{table}

Because the residual under an evaluation mixture is the corresponding weighted average of the per-depth
residuals, the small error under the original mixture partly reflects cancellation across depths;
concentrating the weight on one depth removes that cancellation. Full results for each held-out seed and
evaluation mixture, together with the reverse model direction, are reported in the Supplementary Material.

\subsection{Observable Task Features on GSM8K}
\label{sec:gsm8k_features}
\label{sec:gsm8k}

The synthetic analysis can condition on depth because the generator supplies it. GSM8K~\citep{cobbe2021training}
has no analogous label, so the next question is whether observable features of the problem statement
can support a weaker version of the same prediction checks. The accounting identity still holds on the
observed sample by construction; what is uncertain is whether the available features separate groups
with meaningfully different rates. At the present sample sizes, they do so only weakly.

Both the task setting and the operation change: we move from helper conditioning on the synthetic
tasks to selection from a fixed candidate set on GSM8K. Llama~3.2 serves as the solver and produces
$K$ candidates; Qwen~2.5 serves as the judge and selects one. Here $E_0$
records success of the baseline candidate and $E_1$ success of the selected candidate under the same
exact-match rule. All results in this subsection use $K=2$ on a fixed judged pool of 1319 items;
candidates are generated at temperature $0.7$ and the judge call is deterministic, so variability
comes from the candidate set rather than from the judge. Presentation effects at $K=4$ are audited
separately in Section~\ref{sec:candidate_order}. Pooled over the judged pool, $p_0=0.7180$ and
$p_1=0.7400$, with $\hat c=0.1193$ and $\hat\gamma=0.0164$.

Binning by question length in characters, the estimated rates do vary across bins, so a pooled fit is
in principle composition-dependent. At roughly 260 items per bin, the largest standardized bin residual
against the pooled prediction is
\(0.51\) in absolute value, and the residual root mean square is \(0.008\), providing little evidence of
resolvable bin-specific deviation at the available sample sizes. Under the fitted length-bin rates,
deterministic short-to-long and long-to-short reweighting changes
the pooled prediction by about $0.47$ percentage points. This deterministic plug-in discrepancy does
not shrink with pool size, but at the present bin sizes it is smaller than a binomial
noise scale of roughly $2$--$3$ percentage points. Under random splits of the same
pool, which match the bin composition in expectation, held-out prediction error is roughly
$0.009$--$0.016$ across training fractions.

Question length is one of several readily observable features of the problem statement that we
examined. The question-length bin definitions are reported in the Supplementary Material. Section~\ref{sec:selective_application} uses the feature that
produced the strongest observed variation. The broader point is that the same checks remain available
on a natural benchmark, while their power depends on how much of the underlying variation an
observable feature captures.

\subsection{Selective Application Using Estimated Rates}
\label{sec:selective_application}

The preceding analyses use the rates to predict what an operation will do on a new sample or within
a new group. The same accounting can also inform an application decision by comparing the estimated
correction gain with the estimated corruption loss. In the synthetic holdout for helper-transcript
conditioning in the direction mistral$\to$llama3.2, rates fitted on two seeds predict a loss at every
depth on the third, so the operation is not applied in any depth group. Equation~\eqref{eq:paired_accounting}
writes the two contributions as \((1-p_0)c\) and \(p_0\gamma\), so a baseline estimate $\hat p_0$ and
calibrated rates imply the predicted change
\[
\widehat{\Delta p}(\hat p_0) = (1-\hat p_0)\hat c - \hat p_0 \hat\gamma.
\]
In the reported experiment we apply the operation only when this quantity is positive; application
cost is not modeled. We fit \(\bigl(\hat c(d),\hat\gamma(d)\bigr)\) on two seeds and evaluate the
displayed rule on the third using the held-out \(\hat p_0(d)\). Averaged over the three choices of
held-out seed, always applying the operation yields mean accuracy $0.6487$ against a baseline of
$0.8493$, so the rule avoids a mean loss of $0.2007$. This is correct suppression of a harmful operation
rather than selective routing among depth groups: the rule never turns the operation on, and no held-out post-operation outcomes enter the
group decisions.

On GSM8K, we use the same fixed judged pool to compare six observable features and examine a group-level
application rule using the feature with the strongest observed discrimination. The six features are computed
directly from the problem statement: character length, word count, total number count, distinct-number count,
sentence count, and a structural-keyword score counting the presence of five categories of terms associated
with rates, percentages or financial quantities, comparisons, ratios, and time units. Sentence count is
displayed in Table~\ref{tab:gsm8k_sentcnt_gating} because it produced the strongest observed discrimination
among these features, a selection made after examining all six rather than specified in advance. The rule
applies the operation in the three lower sentence-count groups and does not apply it in the highest group
(6--13 sentences; $N=134$, $\hat c=0.040$, $\hat\gamma=0.075$, predicted change $-0.022$), and all four
predicted signs agree with the observed sign of $p_1-p_0$, though within this pool the predicted change is the smoothed counterpart of the observed change, so the agreement is a consistency check rather than held-out validation. The predicted margins are small, the groups are
unequal in size, and the decisions are group-level; question length, by contrast, produced no detectable
discrimination in the same comparison.

\begin{table}[htbp]
\centering
\caption{\textbf{Application decisions from estimated correction and corruption rates.}
Panel A reports the leave-one-seed-out synthetic result for
mistral$\to$llama3.2: the predicted change is nonpositive in all five depth
groups, so the operation is not applied in any group and the loss from always
applying it is avoided. Panel B reports the exploratory GSM8K sentence-count
grouping at $K=2$. Six features computed directly from the problem statement
were examined, and sentence count is displayed because it produced the
strongest observed discrimination; this
choice was made after examining all six. The decisions are group-level, the predicted margins are
modest, and the feature comparison, sentence-count selection, and direction check all use the same fixed
judged pool. The final column reports whether the sign of the predicted change matches the sign of the
observed change $p_1-p_0$. Estimates of $c$ and $\gamma$ are
Jeffreys-smoothed; $p_0$ and $p_1$ are empirical accuracies.}
\label{tab:gsm8k_sentcnt_gating}
\small

\textbf{Panel A: held-out synthetic decision}

\begin{adjustbox}{max width=\linewidth}
\begin{tabular}{@{}lccrrrr@{}}
\toprule
Scope & Decision & Groups not applied & Baseline & Always apply & Apply by rule & Avoided loss \\
\midrule
Five depth groups & Do not apply & 5/5 & 0.8493 & 0.6487 & 0.8493 & 0.2007 \\
\bottomrule
\end{tabular}
\end{adjustbox}

\medskip
\textbf{Panel B: exploratory GSM8K sentence-count groups}

\begin{adjustbox}{max width=\linewidth}
\begin{tabular}{@{}lrrrrrrcc@{}}
\toprule
Sentence count & $N$ & $p_0$ & $p_1$ & $\hat c$ & $\hat\gamma$
& Predicted change & Decision & Signs agree \\
\midrule
1--3  & 682 & 0.7771 & 0.8035 & 0.1405 & 0.0066 & $+0.026$ & Apply        & yes \\
4     & 342 & 0.7135 & 0.7310 & 0.1263 & 0.0265 & $+0.017$ & Apply        & yes \\
5     & 161 & 0.6273 & 0.6770 & 0.1557 & 0.0147 & $+0.049$ & Apply        & yes \\
6--13 & 134 & 0.5373 & 0.5149 & 0.0397 & 0.0753 & $-0.022$ & Do not apply & yes \\
\bottomrule
\end{tabular}
\end{adjustbox}
\end{table}

Taken together the two examples mark the range of what measured rates buy. In the controlled setting,
where the grouping variable is supplied and the operation is clearly harmful, the decision is correct
and its value is large. On the natural benchmark, where groups are approximated by a feature chosen
after inspection, the predicted and observed signs agree but the margins are modest and the evidence
is exploratory.
\section{Effects of Information Supplied to Protocol Steps}
\label{sec:information_supplied}

Section~\ref{sec:prediction} examined whether estimated rates support prediction and application. We
now change information supplied to an otherwise matched operation. In the first comparison, the same realized candidate set is presented in a different order. In the second, the same
task and initially passing program are presented to a revision call with or without a helper artifact. These
paired comparisons show how presentation and added context alter correction and corruption.

\subsection{Candidate Order and Selection}
\label{sec:candidate_order}
\label{sec:gsm8k_posbias}

We return to the GSM8K selection setup of Section~\ref{sec:gsm8k_features}: Llama~3.2 generates the
candidate set and Qwen~2.5 selects one candidate. Here we use a separate 200-item GSM8K audit subset
with \(K=4\); candidate~1 is the baseline answer, which we call the anchor. The standard condition
presents the anchor first. A deterministic per-item permutation, \texttt{posshuffle}, changes only the
presentation order before the same judge call; after selection, the chosen position is mapped back to the
original candidate index. The realized candidates, judge model, instruction, and decoding are unchanged.
Here \(E_0\) records success of the anchor and \(E_1\) success of the selected candidate.

Panel A of Table~\ref{tab:gsm8k_order_and_structure} separates selection behavior from final success. Under
standard ordering, the judge chooses both the first presented candidate and the original anchor with
probability \(0.790\). After reshuffling, it still chooses the first presented candidate with probability
\(0.675\), but chooses the original anchor only \(0.280\) of the time, close to the uniform reference
\(1/K=0.25\). Anchor retention also becomes nearly independent of its initial status: the probabilities
\(0.926\) and \(0.404\), conditional on \(E_0=1\) and \(E_0=0\), become \(0.284\) and \(0.269\).
Moving away from the anchor raises both the correction estimate, from \(0.255\) to \(0.349\), and the
corruption estimate, from \(0.017\) to \(0.084\); selected-answer accuracy changes from \(0.795\) to
\(0.770\), a five-item difference that is small relative to sampling variation at \(N=200\). The larger and
more diagnostic change is in which candidate the judge selects. Standard-order performance therefore partly
reflects preservation of the anchor, especially when the anchor is successful.

Whether that preservation helps depends on the realized candidate set. Panel B partitions the same 200
items into three groups. In the no-diversity group, all four candidates reduce to the same normalized answer,
so reshuffling leaves selected-answer accuracy unchanged at \(0.967\). In the
diversity-without-headroom group (\(N=81\)), candidates differ but no alternative improves on the anchor;
reshuffling lowers accuracy from \(0.716\) to \(0.593\), while \(\hat\gamma\) rises from \(0.041\) to
\(0.205\). In the smaller headroom group (\(N=28\)), the anchor is wrong and at least one alternative is
correct; reshuffling raises accuracy from \(0.464\) to \(0.643\), corresponding to five additional successful
selections, while \(\hat c\) rises from \(0.466\) to \(0.638\). In this audit, the observed order effect is
therefore not uniformly harmful or benign: its direction differs with the rescue opportunity present in the
realized candidate set.
\begin{table}[htbp]
\centering
\caption{\textbf{Candidate order and candidate-set structure on the fixed GSM8K \(K=4\) audit
subset.} Panel A compares standard ordering with deterministic per-item reshuffling on the same 200
realized candidate sets. \(p_{\mathrm{oracle}}\) is the empirical best achievable accuracy on those sets;
\(P(\mathrm{first})\) is selection of the first presented candidate, and \(P(\mathrm{anchor})\) is selection
of original candidate~1 after shuffled positions are mapped back to original indices. Panel B partitions the
same items by candidate-set structure. Rates \(\hat c\) and \(\hat\gamma\) are Jeffreys-smoothed; a dash
indicates that corruption is undefined because the group contains no successful anchors. In the no-diversity
group, no item changes status: \(\hat c=0.125\) and \(\hat\gamma=0.006\) are the smoothed values obtained
from zero observed corrections among three unsuccessful anchors and zero observed corruptions among 88
successful anchors.}
\label{tab:gsm8k_order_and_structure}
\label{tab:gsm8k_posbias_k4}
\label{tab:gsm8k_judgek_k4_by_stratum}
\scriptsize
\setlength{\tabcolsep}{3.2pt}

\textbf{Panel A: presentation-order comparison}

\begin{adjustbox}{max width=\linewidth}
\begin{tabular}{@{}lrrrrrr@{}}
\toprule
Setting & \(N\) & \(p_0\) & \(p_1\) & \(p_{\mathrm{oracle}}\) &
\(\hat c\) & \(\hat\gamma\) \\
\midrule
Standard order & 200 & 0.740 & 0.795 & 0.880 & 0.255 & 0.017 \\
\texttt{posshuffle} & 200 & 0.740 & 0.770 & 0.880 & 0.349 & 0.084 \\
\bottomrule
\end{tabular}
\end{adjustbox}

\smallskip
\begin{adjustbox}{max width=\linewidth}
\begin{tabular}{@{}lrrrr@{}}
\toprule
Setting & \(P(\mathrm{first})\) & \(P(\mathrm{anchor})\) &
\(P(\mathrm{anchor}\mid E_0=1)\) & \(P(\mathrm{anchor}\mid E_0=0)\) \\
\midrule
Standard order & 0.790 & 0.790 & 0.926 & 0.404 \\
\texttt{posshuffle} & 0.675 & 0.280 & 0.284 & 0.269 \\
\bottomrule
\end{tabular}
\end{adjustbox}

\medskip
\textbf{Panel B: comparison by candidate-set structure}

\begin{adjustbox}{max width=\linewidth}
\begin{tabular}{@{}llrrrrrr@{}}
\toprule
Candidate-set group & Setting & \(N\) & \(p_0\) & \(p_1\) &
\(p_{\mathrm{oracle}}\) & \(\hat c\) & \(\hat\gamma\) \\
\midrule
No diversity & Standard order & 91 & 0.967 & 0.967 & 0.967 & 0.125 & 0.006 \\
No diversity & \texttt{posshuffle} & 91 & 0.967 & 0.967 & 0.967 & 0.125 & 0.006 \\
Diversity, no headroom & Standard order & 81 & 0.741 & 0.716 & 0.741 & 0.023 & 0.041 \\
Diversity, no headroom & \texttt{posshuffle} & 81 & 0.741 & 0.593 & 0.741 & 0.023 & 0.205 \\
Headroom & Standard order & 28 & 0.000 & 0.464 & 1.000 & 0.466 & -- \\
Headroom & \texttt{posshuffle} & 28 & 0.000 & 0.643 & 1.000 & 0.638 & -- \\
\bottomrule
\end{tabular}
\end{adjustbox}
\end{table}

\subsection{Helper Context in the MBPP Audit}
\label{sec:mbpp_helper_context}
\label{sec:coding_check}

The second comparison changes the presence of helper context rather than candidate order. We use 960
executable Python tasks derived from the Mostly Basic Python Problems benchmark
\citep{austin2021program}. Llama~3.2 first produces one program for each task. Unlike the
helper-conditioning experiment of Section~\ref{sec:prediction}, where the target re-solves the task
conditioned on a separately produced worked solution, the MBPP comparison then asks Llama~3.2 to revise
its own existing program. The audit conditions on the 357 programs that pass the unit tests and applies
this Llama~3.2 revision call under two input configurations. In the draft-only configuration, the call
receives the task and Llama~3.2's existing passing program. In the helper-augmented configuration, it
receives those same inputs together with a Mistral-generated helper artifact. The decoding settings, core
instruction, task, initial program, and tests remain fixed.

Because every program in this slice passes before revision, the comparison is a one-sided corruption audit:
it measures loss of pass status and does not estimate correction of initially failing programs. As
Table~\ref{tab:mbpp_helper_corruption} shows, the draft-only operation turns \(28\) of \(357\) passing
programs into failures, a corruption rate of \(0.078\). Adding the helper artifact raises the count to
\(100\) and the rate to \(0.280\), a paired increase of \(0.202\) with bootstrap 95\% confidence interval
\([0.154,0.249]\). The paired discordances point in the same direction: 83 programs fail only when the
helper is supplied, whereas 11 fail only in the draft-only configuration.

The result identifies a concrete cost of additional context in this revision operation. Because every initial
program in the selected slice passes, the lower final pass rate is entirely attributable to corruption. The
design problem created by the helper artifact is therefore preservation of already-passing programs while
retaining any benefit the additional context may offer elsewhere. Prompt templates, row-level examples, and
reproduction details are provided in the Supplementary Material
(Appendix~\suppref{sec:coding_check_supp}) and the accompanying coding-check notebook.
\begin{table}[htbp]
\centering
\caption{\textbf{One-sided MBPP corruption audit on 357 initially passing programs.}
The two configurations use the same tasks, initial programs, downstream revision model, decoding settings,
core instruction, and unit tests; the helper-augmented configuration additionally supplies a
Mistral-generated artifact. The confidence interval is the bootstrap interval for the paired increase in
corruption rate. ``Helper-only failure'' and ``draft-only failure'' are the discordant paired outcomes.}
\label{tab:mbpp_helper_corruption}
\small
\begin{adjustbox}{max width=\linewidth}
\begin{tabular}{@{}lrrl@{}}
\toprule
Quantity & Count & Rate & 95\% CI \\
\midrule
Draft-only corruption & 28 & 0.078 & -- \\
Helper-augmented corruption & 100 & 0.280 & -- \\
Paired increase & \(+72\) & \(+0.202\) & \([0.154,0.249]\) \\
\midrule
Helper-only failure & 83 & -- & -- \\
Draft-only failure & 11 & -- & -- \\
\bottomrule
\end{tabular}
\end{adjustbox}
\end{table}
\section{Combining Measurements across Successive Steps}
\label{sec:successive_steps}

The preceding sections considered one operation at a time. A protocol stack raises a further question:
whether measurements of adjacent operations can be combined to predict the result after both have run.
That combination is useful only when the intermediate outcome retains the information needed by the
later operation. This section makes that requirement observable, first within each generation seed and
then by estimating adjacent transitions on two seeds and comparing them with direct outcomes on a held-out
third.

\subsection{Intermediate Success as a State Summary}
\label{sec:intermediate_correctness}
\label{sec:composition}
\label{sec:composition_logging}

We study the same synthetic corpus defined in Section~\ref{sec:prediction}, using all six groups and
600 items per seed. The stack has three checkpoints. \(E_0\) records success of Mistral's independent
answer. \(E_1\) records success of Mistral's answer after conditioning on a Llama~3.2 helper
transcript. Qwen~2.5 then performs verify-and-fix: it retains the incoming answer when it accepts it
and produces a replacement answer when it judges it wrong. \(E_2\) records success of that final
answer. Unlike the verdict-only diagnostic discussed in Section~\ref{sec:paired_scope}, this second
operation transforms or retains the answer rather than returning only a classification verdict.

The two adjacent transitions tell us how success changes from \(E_0\) to \(E_1\) and from \(E_1\)
to \(E_2\). Combining them assumes that two items with the same value of \(E_1\) are equivalent for
the second operation, even if they reached that state through different histories. To test that
assumption, for \(a<b\) let
\[
T_{ab}(i,k)=\Pr(E_b=k\mid E_a=i),\qquad i,k\in\{0,1\}.
\]
We estimate the adjacent transitions \(T_{01}\) and \(T_{12}\), form their product
\begin{equation}
\label{eq:composition_kernel}
T_{02}^{\mathrm{comp}}=T_{01}T_{12},
\end{equation}
and compare it with the direct transition \(T_{02}^{\mathrm{direct}}\) estimated from the paired
\((E_0,E_2)\) outcomes. If the second operation depends on earlier history only through \(E_1\),
then \(E_2\perp E_0\mid E_1\), which implies
\(T_{02}^{\mathrm{direct}}=T_{02}^{\mathrm{comp}}\). The converse need not hold: history-dependent
differences can cancel when the outcomes are collapsed to two rows. The comparison therefore tests
how well the intermediate success bit summarizes this stack; equality of the collapsed transitions is
not itself a proof of conditional independence.

We summarize the discrepancy in two ways. The row-averaged \(\ell_1\) gap,
\(\Delta_{\mathrm{comp}}^{\mathrm{mean}}\), measures the overall discrepancy between the direct and
composed rows. The maximum entrywise gap, \(\Delta_{\mathrm{comp}}^{\max}\), records the largest
single difference. Full formulas and support-sensitive variants are given in the Supplementary
Material, Sec.~\suppref{app:composition_gap}.

\subsection{Composition within a Generation Seed}
\label{sec:matched_composition}
\label{sec:composition_diagnostics}

We first estimate the direct and adjacent transitions separately within each generation seed. This is
a within-seed comparison: for seed 123, for example, all three transitions are estimated from the same
600 tasks generated under seed 123, and the calculation is repeated independently for seeds 124 and
125.

Propagating the observed initial success proportion through adjacent transitions estimated on the same
items gives a pooled final-accuracy residual of \(0.0002\). This is a useful implementation consistency
check, but with unsmoothed empirical transitions the corresponding marginal identity holds by
construction; the small nonzero value arises from smoothing and is not the evidence for the state
summary. That evidence comes from comparing the direct and composed transitions themselves.

Panel A of Table~\ref{tab:composition_summary} aggregates the per-group diagnostics defined above. Its
weighted mean averages \(\Delta_{\mathrm{comp}}^{\mathrm{mean}}\) over the six groups, and its maximum
is the largest \(\Delta_{\mathrm{comp}}^{\max}\) across them. The weighted means are \(0.0233\),
\(0.0592\), and \(0.0478\) for seeds 123, 124, and 125, and the maximum reaches \(0.1313\). Thus,
within each seed, the intermediate success bit gives a close average summary but not uniform agreement
across groups.

\subsection{Held-Out Composition and Conditional Support}
\label{sec:heldout_composition}

For the held-out test, we estimate \(T_{01}\) and \(T_{12}\) on two seeds and compare their product
with the direct \(T_{02}\) observed on the third. Panel B differs from Panel A in two respects: the
adjacent transitions are estimated on calibration seeds and evaluated against a held-out direct transition,
and the six task groups are pooled rather than summarized separately before
\(\Delta_{\mathrm{comp}}^{\mathrm{mean}}\) is computed. The pooled leave-one-seed-out composition gaps
are \(0.1031\), \(0.1667\), and \(0.1227\) for held-out seeds 123, 124, and 125. When the two
initial-success rows are instead weighted by their held-out support, the corresponding gaps are \(0.0277\),
\(0.0200\), and \(0.0301\) (Panel B).

The same adjacent-transition estimates can also be used to predict the final success proportion in each
task group. Panel C reports this separate accuracy-level check for seed 125, using seeds 123 and 124 for
calibration. A single pooled composed transition misses the \(2\times2\) systems by \(-0.1209\) and
depth 5 by \(-0.1073\). Group-conditioned composition reduces those residuals to \(-0.0169\) and
\(-0.0623\), respectively, and reduces absolute error in all six groups. The improvement is smallest
at depth 4, where the residual changes from \(0.0370\) to \(0.0327\), so conditioning improves rather
than eliminates the remaining error.

Taken together, the three panels separate within-seed composition, pooled held-out comparison, and
group-level final-accuracy prediction. Within a seed, the binary success state gives a close average summary
across groups; that adequacy does not by itself establish that adjacent transitions estimated on calibration
seeds will match the direct transition observed on a held-out seed, which Panel B evaluates directly. The
smaller support-weighted held-out gaps, together with the supplementary support diagnostics, are consistent
with greater finite-sample sensitivity when one held-out \(E_0\) row is thin; the remaining gaps may also
reflect structural mismatch. Full by-group, support, convergence, and signed-gap diagnostics are reported
in the Supplementary Material.

\begin{table}[htbp]
\centering
\caption{\textbf{Within-seed and held-out composition diagnostics.}
Panel A compares the direct two-step transition with the product of the adjacent transitions when all
quantities are estimated within the same generation seed. The weighted mean averages the row-averaged
gap over the six task groups; the maximum records the largest binwise entrywise gap. Panel B estimates
the adjacent transitions on two seeds and evaluates the direct-versus-composed gap on the third.
The row-weighted value weights the two \(E_0\) rows by their held-out support. Panel C is a distinct
final-accuracy prediction check for held-out seed 125: residuals are empirical minus predicted
\(p_2\), and the group-conditioned prediction uses a separate composed transition for each of the five
arithmetic depths and the \(2\times2\) family. All transition estimates are Jeffreys-smoothed.}
\label{tab:composition_summary}
\label{tab:comp_gap_pooled}
\label{tab:composition_seed_holdout}
\small

\textbf{Panel A: within-seed transition gaps}

\begin{adjustbox}{max width=\linewidth}
\begin{tabular}{@{}lrrr@{}}
\toprule
Generation seed & \(N\) & Weighted mean gap & Maximum binwise gap \\
\midrule
123 & 600 & 0.0233 & 0.0328 \\
124 & 600 & 0.0592 & 0.1313 \\
125 & 600 & 0.0478 & 0.0956 \\
\bottomrule
\end{tabular}
\end{adjustbox}

\medskip
\textbf{Panel B: leave-one-seed-out transition gaps}

\begin{adjustbox}{max width=\linewidth}
\begin{tabular}{@{}lrr@{}}
\toprule
Held-out seed & Pooled gap & Row-weighted gap \\
\midrule
123 & 0.1031 & 0.0277 \\
124 & 0.1667 & 0.0200 \\
125 & 0.1227 & 0.0301 \\
\bottomrule
\end{tabular}
\end{adjustbox}

\medskip
\textbf{Panel C: held-out final-accuracy residuals for seed 125}

\begin{adjustbox}{max width=\linewidth}
\begin{tabular}{@{}lrrr@{}}
\toprule
Task group & \(N\) & Pooled residual & Group-conditioned residual \\
\midrule
\(2\times2\) systems & 100 & \(-0.1209\) & \(-0.0169\) \\
Depth 1 & 100 & \( 0.0366\) & \( 0.0193\) \\
Depth 2 & 100 & \( 0.0653\) & \(-0.0187\) \\
Depth 3 & 100 & \( 0.0449\) & \( 0.0118\) \\
Depth 4 & 100 & \( 0.0370\) & \( 0.0327\) \\
Depth 5 & 100 & \(-0.1073\) & \(-0.0623\) \\
\bottomrule
\end{tabular}
\end{adjustbox}
\end{table}
\FloatBarrier
\section{Discussion and Limitations}
\label{sec:discussion}

The paired audit separates retrospective accounting from prospective reuse. On an observed set of tasks,
the four paired counts determine exactly how many outputs were corrected and corrupted, and
Equation~\eqref{eq:paired_accounting} accounts for the net change. Reusing estimated rates beyond that
set is an empirical question. In Section~\ref{sec:structural_depth}, calibration and held-out samples come
from the same generator-induced task distribution; Section~\ref{sec:population_composition} then changes
the evaluation task distribution by reweighting the generator-defined depth groups while keeping the fitted
rates and observed outcomes fixed. Sections~\ref{sec:information_supplied} and~\ref{sec:successive_steps}
test two further requirements for reuse: sensitivity to the information supplied to an operation and
composition across successive operations.

The synthetic experiments show why task composition matters. Under the original equal-depth mixture,
pooled and depth-conditioned predictions are similar, but across the single-depth reweightings conditioning
on generator-supplied depth roughly halves average prediction error, from \(0.0711\) to \(0.0375\).
Correction and corruption rates differ across these generator-defined groups, so changing their weights
changes the aggregate behavior represented by a pooled rate. Conditioning also divides the observations
available for estimating each conditional rate and therefore requires adequate support. More generally,
changes in the composition of an evaluation task distribution are a form of dataset shift
\citep{quinonero2009datasetshift,sugiyama2012covariateshift}.

GSM8K illustrates the harder case in which no analogous generator-supplied grouping variable is available.
Readily observable problem features separate the flows much less clearly at the available sample sizes.
For question-length bins, deterministic reweighting changes the pooled prediction by about \(0.47\)
percentage points. This plug-in discrepancy does not shrink with pool size, but at the present bin sizes it
is smaller than the roughly \(2\)--\(3\) percentage-point binomial scale. The sentence-count application rule is exploratory because the feature was selected after examining six candidate features on the same judged pool. Thus conditioning is useful only when available features
resolve meaningful variation; weak observed variation does not make a pooled rate a population-independent
property of a model.

The candidate-order audit reveals a different dependence. Under standard ordering, first position and the
original anchor coincide, and each is selected with probability \(0.790\). After reshuffling, the first
presented candidate is still selected with probability \(0.675\), while selection of the original anchor
falls to \(0.280\); the strong difference in anchor retention between initially successful and unsuccessful
items disappears. The accuracy consequence depends on the realized candidate set. When diverse alternatives
offer no headroom, reshuffling lowers accuracy from \(0.716\) to \(0.593\); when the anchor is wrong and a
correct alternative is available, it raises accuracy from \(0.464\) to \(0.643\). The aggregate \(0.795\)
versus \(0.770\) difference over 200 items is small relative to sampling variation, but the selection changes
are substantial. Reporting \(p_{\mathrm{oracle}}\) alongside \(p_0\) and \(p_1\) separates whether a
realized candidate set contains a correct alternative from whether the judge selects one. Related positional
and presentation effects in model-based judging have been reported by
\citet{zheng2023mtbench} and \citet{wang2023large}.

MBPP deliberately tests the same measurement outside exact-scored arithmetic: success is passage of
executable unit tests rather than numeric exact match. Under this different domain and success criterion,
adding the Mistral-generated helper artifact raises corruption among initially passing programs from
\(0.078\) to \(0.280\), a paired increase of \(0.202\) with bootstrap 95\% CI
\([0.154,0.249]\). Because the audit conditions on initially passing programs, it does not measure whether
the same context repairs initially failing ones. The experiment therefore extends the corruption measurement
beyond exact arithmetic while isolating a substantial cost of added context.

The decomposition can also support an application decision. In the synthetic
\texttt{mistral}$\to$\texttt{llama3.2} holdout, calibrated rates predict a loss in every depth group, so
suppressing helper-transcript conditioning avoids the mean accuracy loss of \(0.2007\) incurred by always
applying it. Here the grouping variable is supplied by the generator and the decision is evaluated on
held-out seeds. The GSM8K rule is weaker because its feature was selected after inspection, its predicted
margins are modest, and the decisions are group-level. A deployment rule would additionally need to account
for call cost and be validated on data not used to select its features or thresholds.

Composition is the form of reuse most directly relevant to multi-step protocols and agents: whether
measurements of local operations can be combined to predict a longer workflow
\citep{dohan2022language}. In our stack, the retained state is deliberately compressed to the binary
success bit, so its adequacy can be tested directly. Within a seed, weighted mean direct-versus-composed
gaps range from \(0.0233\) to \(0.0592\), although the maximum groupwise gap reaches \(0.1313\), so
agreement is not uniform across groups. In held-out comparisons, pooled gaps of \(0.1031\)--\(0.1667\)
fall to \(0.0200\)--\(0.0301\) after weighting the initial-success rows by their held-out support;
separately, task-group conditioning reduces final-accuracy prediction error in all six groups for held-out
seed 125. Intermediate success is therefore a useful compact state for this stack, while the held-out checks
show why local measurements should not simply be assumed to compose in an agent or longer protocol. A
persistent, well-supported discrepancy could indicate missing predictive history, although finite-sample
support and calibration-to-target mismatch can also contribute.

Several limits bound the direct empirical scope. Our experiments use a small set of open-weight checkpoints
on controlled mathematics, GSM8K, and MBPP. Independent work by \citet{liu2026feedbackcontrol}, including
experiments with proprietary systems not tested here, reports both correction and error introduction across
additional models and task settings, providing evidence that the distinction is not peculiar to these
checkpoints. Our prediction, input-sensitivity, selective-application, and composition results remain direct
evidence only for the settings studied here. Most outcomes use exact match, while MBPP deliberately uses
executable tests; graded or non-unique outputs would require an auditable success rule. The MBPP audit
measures corruption only, the \(K=4\) presentation audit contains 200 items, and the GSM8K feature analysis
is exploratory. The synthetic operations use deterministic decoding, whereas GSM8K candidate generation
includes sampling variation. Longer adaptive workflows, repeated stochastic calls, substantially different
capability levels, and richer forms of feedback remain outside the direct evidence. Paired outcomes establish
what changed, not why a model revised or preserved a particular answer.

The practical contribution is therefore a measurement discipline rather than a universal performance law.
For each auditable operation, record paired outcomes and report counts together with correction and
corruption rates. Before reusing those rates, test them on held-out data representative of the intended task
distribution and examine task groups whose relative weights may change. When supplied information changes,
remeasure the flows. For a multi-step protocol or agent, compare predictions from adjacent transitions with
direct end-to-end outcomes before assuming composition. When these checks succeed, correction and
corruption provide a compact description that can guide protocol design; when they fail, the pattern can
help localize the obstacle to population composition, sensitivity to supplied information, or loss of
predictive history between steps.

\section{Conclusion}
\label{sec:conclusion}

An added model call can both correct initially unsuccessful outputs and corrupt initially successful ones.
Aggregate accuracy records only the net result. The paired audit developed here separates those two flows
for a specified operation, population, input configuration, and success rule. Their weighted difference exactly
recovers the observed change in success, but the practical value of the decomposition lies in the empirical
questions it makes testable: whether measured rates predict new cases, how they change when the operation's
inputs change, and whether adjacent-step measurements combine across a stack.

Across the experiments, those questions have different answers in different settings. Calibration estimates
track held-out outcomes in the matched synthetic setting, but pooled estimates can fail under changes in
population composition; conditioning on generator-supplied structural depth improves prediction on average
under controlled reweighting.
On GSM8K, observable features separate groups only weakly at the available sample sizes, and the resulting
application rule is exploratory. Candidate order
and helper context can materially alter correction and corruption even when other elements of the operation
remain fixed. In the two-step stack, intermediate success provides a useful average summary within a
generation seed, yet discrepancies remain across task groups and when adjacent transitions estimated on calibration seeds are compared with the direct transition on a held-out seed. These results support correction and corruption as a compact audit of a specified protocol step,
not as invariant properties of a model or a universal state representation.

Independent work by \citet{liu2026feedbackcontrol} defines error-correction and error-introduction rates
that coincide exactly with \(c\) and \(\gamma\) and, in its repeated self-revision setting, independently
derives the same accuracy accounting identity and break-even boundary, together with its consequence
that the break-even requirement rises sharply with baseline accuracy. The prediction, input-sensitivity, and composition results reported here are separate empirical contributions.

Extending the approach to longer or more adaptive protocols will require testing whether the retained state is
adequate for each downstream operation. Confidence, error type, or transcript features are illustrative
possibilities, not states identified as sufficient by our experiments. The central methodological
recommendation is therefore simple: log paired outcomes at each auditable step, report correction and
corruption separately, and validate any reuse, redesign, or composition in the population where the protocol
will operate.

\bibliographystyle{tmlr}
\bibliography{references}

\end{document}